\documentclass[11pt]{article}
\usepackage{colacl}
\usepackage{graphicx}
\def\sref#1{\S\ref{#1}}
\def\fref#1{fig.~\ref{#1}}

\title{Approximation and Exactness in Finite State Optimality Theory}
\author{Dale Gerdemann                    \\
        University of T\"ubingen          \\
        {\tt dg@sfs.nphil.uni-tuebingen.de}
        \And Gertjan van Noord            \\
        University of Groningen           \\
        {\tt vannoord@let.rug.nl}          }
\begin{document}
\thispagestyle{plain}
\maketitle
\begin{abstract}
  Previous work \cite{frank-satta,karttunen:98} has shown that
  Optimality Theory with {\em gradient} constraints generally is not
  finite state.  A new finite-state treatment of gradient constraints
  is presented which improves upon the approximation of
  \newcite{karttunen:98}. The method turns out to be exact, and very
  compact, for the syllabification analysis of
  \newcite{prince-smolensky:93}.
\end{abstract}

\section{\label{intro}Introduction}

Finite state methods have proven quite successful for encoding
rule-based generative phonology \cite{john:form72,kapl:regu94}.
Recently, however, Optimality Theory \cite{prince-smolensky:93} has
emphasized phonological accounts with default constraints on surface
forms. While Optimality Theory (OT) has been successful in explaining
certain phonological phenomena such as {\em conspiracies}
\cite{kisseberth}, it has been less successful for computation. The
negative result of \newcite{frank-satta} has shown that in the general
case the method of counting constraint violations takes OT beyond the
power of regular relations. To handle such constraints,
\newcite{karttunen:98} has proposed a finite-state approximation that
counts constraint violations up to a predetermined bound. Unlike
previous approaches \cite{ellison,walther}, Karttunen's
approach is encoded entirely in the finite state calculus, with no
extra-logical procedures for counting constraint violations. 

In this paper, we will present a new approach that seeks to minimize
constraint violations without counting. Rather than counting, our
approach employs a filter based on matching constraint violations
against violations in alternatively derivable strings. As in
Karttunen's counting approach, our approach uses purely finite state
methods without extra-logical procedures. We show that our {\em
matching} approach is superior to the {\em counting} approach for both
size of resulting automata and closeness of approximation. The
matching approach can in fact exactly model many OT analyses where the
counting approach yields only an approximation; yet, the size of the
resulting automaton is typically much smaller.

In this paper we will illustrate the matching approach and compare it
with the counting approach on the basis of the Prince \& Smolensky
syllable structure example \cite{prince-smolensky:93,ellison,tesar},
for each of the different constraint orderings identified in Prince \&
Smolensky.

\section{\label{fsphon}Finite State Phonology}

\subsection{\label{fsc}Finite State Calculus}

Finite state approaches have proven to be very successful for
efficient encoding of phonological rules. In particular, the work of
\newcite{kapl:regu94} has provided a compiler from
classical generative phonology rewriting rules to finite state
transducers. This work has clearly shown how apparently procedural
rules can be recast in a declarative, reversible framework.

In the process of developing their rule compiler, Kaplan \& Kay also
developed a high-level finite state calculus. They argue convincingly
that this calculus provides an appropriate high-level approach for
expressing regular languages and relations. The alternative conception
in term of states and transitions can become unwieldy for all but the
simplest cases.\footnote{Although in some cases such a direct
implementation can be much more efficient
\cite{mohri-sproat:96,wia99}.} 

Kaplan \& Kay's finite state calculus now exists in multiple
implementations,  the most well-known of which is that of
\newcite{kart:regu96}. In this paper, however, we will use the
alternative implementation provided by the FSA Utilities
\cite{fsa-two,noord:fsa6,wia99}. 
The FSA Utilities allows the programmer to introduce new regular
expression operators of arbitrary complexity. This higher-level
interface allows us to express our algorithm more easily. The syntax
of the FSA Utilities calculus is summarized in Table~\ref{notation}.

\begin{table}
\small
\begin{tabular}{cl}
\tt []                  & empty string                               \\
\tt [E1,E2,\dots ,En]    & concatenation of \tt E1\dots En          \\
\tt \verb+{}+           & empty language                              \\
\tt \verb+{+E1,E2,\dots ,En\verb+}+    & union of \tt E1\dots En    \\
\tt (E)                 & grouping for op. precedence                \\
\tt E*                  & Kleene closure                             \\
\tt E+                  & Kleene plus                                \\
\tt E\verb+^+           & optionality                                \\
\tt E1 - E2             & difference                                 \\
\tt \~{}E               & complement                                 \\
\tt \verb+$+ E          & containment                                \\ 
\tt E1 \verb+&+ E2      & intersection                               \\
\tt ?                   & any symbol                                 \\
\tt E1 x E2             & cross-product                              \\
\tt A o B               & composition                                \\
\tt domain(E)           & domain of a transduction                   \\
\tt range(E)            & range of a transduction                    \\
\tt identity(E)         & identity transduction\protect\footnotemark \\
\tt inverse(E)          & inverse transduction                       \\
\end{tabular}
\centering
\caption{\label{notation}Regular expression operators.}
\end{table}

\footnotetext{\label{identity}If an expression for a recognizer occurs
in a context where a transducer is required, the identity operation
will be used implicitly for coercion.}

The finite state calculus has proven to be a very useful tool for the
development of higher-level finite state operators
\cite{kart:95,KempeKarttunen,karttunen:96,eacl99-gerdemann-vannoord}.
An interesting feature of most such operators is that they are
implemented using a generate-and-test paradigm. 
\newcite{karttunen:96}, for example, introduces an algorithm for a
leftmost-longest replacement operator. Somewhat simplified, we may
view this algorithm as having two steps. First, the generator freely
marks up possible replacement sites. Then the tester, which is an
identity transducer, filters out those cases not conforming to the
leftmost-longest strategy. Since the generator and tester are both
implemented as transducers, they can be composed into a single
transducer, which eliminates the inefficiency normally associated with
generate-and-test algorithms.

\subsection{\label{fsot}Finite State Optimality Theory}

\begin{figure}[here]
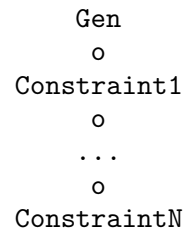

\begin{center}
\begin{tabular}{c}
\tt            Gen       \\
\tt             o        \\
\tt        Constraint1   \\
\tt             o        \\
\tt            ...       \\
\tt             o        \\
\tt        ConstraintN   \\
\end{tabular}
\end{center}
\caption{\label{gen-test}Optimality Theory as Generate and Test}
\end{figure}

The generate-and-test paradigm initially appears to be appropriate for
optimality theory. If, as claimed in \newcite{ellison}, {\em Gen} is
a regular relation and if each constraint can be implemented as an
identity transducer, then optimality theory analyses could be
implemented as in \fref{gen-test}.

The problem with this simple approach is that in OT, a constraint is
allowed to be violated if none of the candidates satisfy that
constraint. \newcite{karttunen:98} treats this problem by providing a
new operator for {\em lenient} composition, which is defined in terms
of the auxiliary operation of priority union. In the FSA Utilities
calculus, these operations can be defined as:\footnote{The notation
{\tt macro(Expr1,Expr2)} is used to indicate that the regular
expression {\tt Expr1} is an abbreviation for the expression {\tt
Expr2}. Because Prolog variables are allowed in both expressions this
turns out to be an intuitive and powerful notation
\cite{wia99}.}

\begin{verbatim}
macro(priority_union(Q,R), 
      {Q, ~domain(Q) o R}).
macro(lenient_composition(S,C), 
      priority_union(S o C, S)).
\end{verbatim}

\noindent The effect here is that the lenient composition of {\tt S}
and {\tt C} is the composition of {\tt S} and {\tt C}, except for
those elements in the domain of {\tt S} that are not mapped to
anything by {\tt S o C}. For these elements not in the domain of {\tt
S o C}, the effect is the same as the effect of {\tt S} alone. 
We use the notation {\tt S lc C} as a succinct notation for the lenient
composition of S and C. Using lenient composition an OT analysis can
be written as in \fref{gen-test-lc}.

\begin{figure}[here]
\begin{center}
\begin{tabular}{c}
\tt          Gen      \\
\tt           lc      \\
\tt      Constraint1  \\
\tt           lc      \\
\tt          ...      \\
\tt           lc      \\
\tt      ConstraintN  \\
\end{tabular}
\end{center}
\caption{\label{gen-test-lc}Optimality Theory as Generate and Test with
  Lenient Composition}
\end{figure}

The use of lenient composition, however, is not sufficient for
implementing optimality theory. In general, a candidate string can
violate a constraint multiple times and candidates that violate the
constraint the least number of times need to be preferred. Lenient
composition is sufficient to prefer a candidate that violates the
constraint 0 times over a candidate that violates the constraint at
least once. However, lenient composition cannot distinguish two
candidates if the first contains one violation, and the second
contains at least two violations. 

The problem of implementing optimality theory becomes considerably
harder when constraint violations need to be counted. As
\newcite{frank-satta} have shown, an OT describes a regular relation
under the assumptions that {\tt Gen} is a regular relation, and each
of the constraints is a regular relation which maps a candidate string
to a natural number (indicating the number of constraint violations in
that candidate), where the range of each constraint is finite. If
constraints are defined in such a way that there is no bound to the
number of constraint violations that can occur in a given string, then
the resulting OT may describe a relation that is not regular. A simple
example of such an OT (attributed to Markus Hiller) is the OT in which
the inputs of interest are of the form {\tt [a*,b*]}, {\em Gen} is
defined as a transducer which maps all {\tt a}'s to {\tt b}'s and all
{\tt b}'s to {\tt a}'s, or alternatively, it performs the identity map
on each {\tt a} and {\tt b}:

\begin{verbatim}
   {[(a x b)*,(b x a)*],     
    [(a x a)*,(b x b)*]}     
\end{verbatim}

\noindent This OT contains only a single constraint, {$^*A$}: a string
should not contain {\tt a}. As can easily be verified, this OT defines
the relation $\{(a^nb^m,a^nb^m) | n\leq m\} \cup \{(a^nb^m,b^na^m) |
m\leq n\}$, which can easily be shown to be non-regular.

Although the OT described above is highly unrealistic for natural
language, one might nevertheless expect that a constraint on syllable
structure in the analysis of Prince \& Smolensky would require an
unbounded amount of counting (since words are of unbounded length),
and that therefore such analyses would not be describable as regular
relations. An important conclusion of this paper is that, contrary to
this potential expectation, such cases in fact {\em can} be shown to
be regular.

\subsection{\label{syl}Syllabification in Finite State OT}

In order to illustrate our approach, we will start with a finite state
implementation of the syllabification analysis as presented in chapter
6 of \newcite{prince-smolensky:93}. This section is heavily based on
\newcite{karttunen:98}, which the reader should consult for more explanation and examples. 

The inputs to the syllabification OT are sequences of consonants and
vowels. The input will be marked up with {\em onset}, {\em nucleus},
{\em coda} and {\em unparsed} brackets; where a syllable is a sequence
of an optional onset, followed by a nucleus, followed by an optional
coda. The input will be marked up as a sequence of such syllables,
where at arbitrary places {\em unparsed} material can intervene. The
assumption is that an unparsed vowel or consonant  is not
spelled out phonetically. Onsets, nuclei and codas are also allowed
to be empty; the phonetic interpretation of such constituents is
{\em epenthesis}. 

First we give a number of simple abbreviations:
\begin{verbatim}
macro(cons,      
      {b,c,d,f,g,h,j,k,l,m,n,
       p,q,r,s,t,v,w,x,y,z}  ).
macro(vowel,   {a,e,o,u,i}).
\end{verbatim}\begin{verbatim}
macro(o_br,    'O['). % onset
macro(n_br,    'N['). % nucleus
macro(d_br,    'D['). % coda
macro(x_br,    'X['). % unparsed
macro(r_br,    ']').
macro(bracket,   
      {o_br,n_br,d_br,x_br,r_br}).
\end{verbatim}\begin{verbatim}
macro(onset,   [o_br,cons^  ,r_br]).
macro(nucleus, [n_br,vowel^ ,r_br]).
macro(coda,    [d_br,cons^  ,r_br]).
macro(unparsed,[x_br,letter ,r_br]).
\end{verbatim}

Following Karttunen, {\em Gen} is formalized as in
\fref{genps}.
\begin{figure*}
\begin{center}
\begin{minipage}{14cm}
\begin{verbatim}
macro(gen,       {cons,vowel}* 
                       o 
                   overparse 
                       o 
                     parse 
                       o 
                syllable_structure ).

macro(parse, replace([[] x {o_br,d_br,x_br},cons, [] x r_br])
                                 o
             replace([[] x {n_br,x_br},     vowel,[] x r_br])).

macro(overparse,intro_each_pos([{o_br,d_br,n_br},r_br]^)).

macro(intro_each_pos(E), [[ [] x E, ?]*,[] x E]).

macro(syllable_structure,ignore([onset^,nucleus,coda^],unparsed)*).
\end{verbatim}
\end{minipage}
\end{center}
\caption{\label{genps}The definition of {\em Gen}}
\end{figure*}
Here, {\tt parse} introduces {\em onset}, {\em coda} or {\em unparsed}
brackets around each consonant, and {\em nucleus} or {\em unparsed}
brackets around each vowel.  The {\tt replace(T,Left,Right)}
transducer applies transducer {\tt T} obligatory within the contexts
specified by {\tt Left} and {\tt Right}
\cite{eacl99-gerdemann-vannoord}. The {\tt replace(T)} transducer is
an abbreviation for {\tt replace(T,[],[])}, i.e. {\tt T} is applied
everywhere. The {\em overparse} transducer introduces optional `empty'
constituents in the input, using the {\em intro\_each\_pos}
operator.\footnote{An alternative would be to define {\it overparse}
with a Kleene star in place of the option operator.  This would
introduce unbounded sequences of empty segments. Even though it can be
shown that, with the constraints assumed here, no optimal candidate
ever contains two empty segments in a row (proposition 4 of
\newcite{prince-smolensky:93}) it is perhaps interesting to note that
defining {\em Gen} in this alternative way causes cases of infinite
ambiguity for the counting approach but is unproblematic for the
matching approach.}

In the definitions for the constraints, we will deviate somewhat from
Karttunen. In his formalization, a constraint simply describes the set
of strings which do not violate that constraint. It turns out to be
easier for our extension of Karttunen's formalization below, as well
as for our alternative approach, if we return to the concept of a
constraint as introduced by Prince and Smolensky where a constraint
adds {\em marks} in the candidate string at the position where the
string violates the constraint. Here we use the symbol {\tt @} to
indicate a constraint violation. After checking each constraint the
markers will be removed, so that markers for one constraint will not
be confused with markers for the next.

\begin{verbatim}
macro(mark_violation(parse),
     replace(([] x @),x_br,[]).

macro(mark_violation(no_coda),
     replace(([] x @),d_br,[]).

macro(mark_violation(fill_nuc),
     replace(([] x @),[n_br,r_br],[])).

macro(mark_violation(fill_ons),
     replace(([] x @),[o_br,r_br],[])).
\end{verbatim}
\begin{verbatim}
macro(mark_violation(have_ons),
     replace(([] x @),[],n_br)
                o
     replace((@ x []),onset,[])).
\end{verbatim}

The {\em parse} constraint simply states that a
candidate must not contain an unparsed constituent. Thus, we add a
mark after each unparsed bracket. The {\em no\_coda} constraint is
similar: each coda bracket will be marked. The {\em fill\_nuc}
constraint is only slightly more complicated: each sequence of a
nucleus bracket immediately followed by a closing bracket is marked.
The {\em fill\_ons} constraint treats empty onsets in the same way.
Finally, the {\em have\_ons} constraint is somewhat more complex. The
constraint requires that each nucleus is preceded by an onset.  This
is achieved by marking all nuclei first, and then removing those marks
where in fact an onset is present.

This completes the building blocks we need for an implementation of
Prince and Smolensky's analysis of syllabification. In the following
sections, we present two alternative implementations which employ
these building blocks. First, we
discuss the approach of \newcite{karttunen:98}, based on the lenient
composition operator. This approach uses a {\em counting} approach for
multiple constraint violations. We will then present an alternative
approach in which constraints eliminate candidates using {\em
  matching}.

\section{\label{count}The Counting Approach}

In the approach of \newcite{karttunen:98}, a candidate set is
leniently composed with the set of strings which satisfy a given
constraint. Since we have defined a constraint as a transducer which
marks candidate strings, we need to alter the definitions somewhat,
but the resulting transducers are equivalent to the transducers
produced by \newcite{karttunen:98}. We use the
(left-associative) {\em optimality operator} {\tt oo} for applying an
OT constraint to a given set of candidates:\footnote{The operators `o' 
and `lc' are assumed to be left associative and have equal
precedence.} 
\bigskip

\begin{verbatim}
macro(Cands oo Constraint,
          Cands
            o
  mark_violation(Constraint)
            lc
        ~ ($ @)
            o
     { @ x [], ? - @}*   ).
\end{verbatim}   
Here, the set of candidates is first composed with the transducer
which marks constraint violations. We then leniently compose the
resulting transducer with \verb+~($ @)+\footnote{As explained in
footnote~\ref{identity}, this will be coerced into an identity
transducer.}, which encodes the requirement that no such marks should
be contained in the string. Finally, the remaining marks (if any) are
removed from the set of surviving candidates.  Using the optimality
operator, we can then combine {\em Gen} and the various constraints as
in the following example (equivalent to figure 14 of
\newcite{karttunen:98}):

\begin{verbatim}
macro(syllabify, gen
                 oo
              have_ons
                 oo
              no_coda
                 oo
              fill_nuc
                 oo
              parse
                 oo
              fill_ons     ).
\end{verbatim}

As mentioned above, a candidate string can violate a
constraint multiple times and candidates that violate the constraint
the least number of times need to be preferred. Lenient composition
cannot distinguish two candidates if the first contains one violation,
and the second contains at least two violations. For example, the
above {\tt syllabify} transducer will assign three outputs to the
input {\tt bebop}:
\begin{verbatim}
O[b]N[e]X[b]X[o]X[p]
O[b]N[e]O[b]N[o]X[p]
X[b]X[e]O[b]N[o]X[p]
\end{verbatim}
In this case, the second output should have been preferred over the
other two, because the second output violates `Parse' only once,
whereas the other outputs violate `Parse' three times.  Karttunen
recognizes this problem and proposes to have a sequence of constraints
Parse0, Parse1, Parse2 \dots ParseN, where each ParseX constraint
requires that candidates not contain more than X unparsed
constituents.\footnote{This construction is similar to the
construction in \newcite{frank-satta}, who used a suggestion in
\newcite{ellison}.} In this case, the resulting transducer only {\em
approximates} the OT analysis, because it turns out that for any X
there are candidate strings that this transducer fails to handle
correctly (assuming that there is no bound on the length of candidate
strings).

Our notation is somewhat different, but equivalent to the notation
used by Karttunen. Instead of a sequence of constraints Cons0 \dots
ConsX, we will write {\tt Cands oo Prec :: Cons}, which is read as:
apply constraint {\tt Cons} to the candidate set {\tt Cands} with {\em
precision} {\tt Prec}, where ``precision'' means the predetermined
bound on counting. For example, a variant of the {\tt syllabify}
constraint can be defined as:
\begin{verbatim}
macro(syllabify, gen
                 oo
              have_ons
                 oo
              no_coda
                 oo
              1 :: fill_nuc
                 oo
              8 :: parse
                 oo
              fill_ons     ).
\end{verbatim}
\noindent Using techniques described in \sref{comparison}, this
variant can be shown to be {\em exact} for all strings of length $\leq 
10$. Note that if no precision is specified, then a precision of 0 is
assumed. 

This construct can be defined as follows (in the
actual implementation the regular expression is computed dynamically
based on the value of {\tt Prec}):

\begin{verbatim}
macro(Cands oo 3 :: Constraint,
          Cands
             o
  mark_violation(Constraint)
            lc
  ~ ([($ @),($ @),($ @),($ @)])
            lc
    ~ ([($ @),($ @),($ @)])
            lc
      ~ ([($ @),($ @)])
            lc
        ~ ($ @)
             o
     { @ : [], ? - @}*  ).
\end{verbatim}

\section{\label{match}The Matching Approach}
\subsection{Introduction}
In order to illustrate the alternative approach, based on matching we
return to the {\tt bebop} example given earlier, repeated here:

{\small\begin{verbatim}
c1:   O[ b ] N[ e ] X[ b ] X[ o ] X[ p ]
c2:   O[ b ] N[ e ] O[ b ] N[ o ] X[ p ]
c3:   X[ b ] X[ e ] O[ b ] N[ o ] X[ p ]
\end{verbatim}}

Here an instance of 'X[' is a constraint violation, so c2 is the best
candidate. By counting, one can see that c2 has one violation, while
c1 and c3 each have 3. By matching, one can see that all candidates
have a violation in position 13, but c1 and c3 also have violations in
positions not corresponding to violations in c2. As long the positions
of violations line up in this manner, it is possible to construct a
finite state filter to rule out candidates with a non-minimal number
of violations. The filter will take the set of candidates, and
subtract from that set all strings that are similar, except that
they contain additional constraint violations.

Given the approach of marking up constraint violations introduced
earlier, it is possible to construct such a matching filter. Consider again
the `bebop' example. If the violations are marked, the candidates of
interest are:

{\small\begin{verbatim}
O[   b ] N[   e ] X[ @ b ] X[ @ o ] X[ @ p ]
O[   b ] N[   e ] O[   b ] N[   o ] X[ @ p ]
X[ @ b ] X[ @ e ] O[   b ] N[   o ] X[ @ p ]
\end{verbatim}}

For the filter, we want to compare alternative mark-ups for the same
input string. Any other differences between the candidates can be 
ignored. So the first step in constructing the filter is to eliminate
everything except the markers and the original input. For the syllable 
structure example, finding the original input is easy since it never
gets changed. For the ``bebop'' example, the filter first constructs:
\begin{verbatim}
    b   e @ b @ o  @  p
    b   e   b   o  @  p
  @ b @ e   b   o  @  p
\end{verbatim}

Since we want to rule out candidates with at least one more constraint
violation than necessary, we apply a transducer to this set which
inserts at least one more marker. This will yield an infinite set of
bad candidates each of which has at least two markers and with one of
the markers coming directly before the final `p'.

In order to use this set of bad candidates as a filter, brackets have
to be reinserted. But since the filter does not care about positions
of brackets, these can be inserted randomly. The result is the set of
all strings with at least two markers, one of the markers coming
directly before the final `p', and arbitrary brackets anywhere. This
set includes the two candidates c1 and c3 above. Therefore, after
applying this filter only the optimal candidate survives. The three
steps of deleting brackets, adding extra markers and randomly
reinserting brackets are encoded in the {\tt add\_violation} macro
given in \fref{addvio}.

\begin{figure*}
\begin{center}
\begin{minipage}{13cm}
\begin{verbatim}
macro(add_violation,
      {(bracket x []), ? - bracket}*    % delete brackets
                     o
           [[? *,([] x @)]+, ? *]       % add at least one @
                     o
      {([] x bracket), ? - bracket}*    % reinsert brackets
     ).
\end{verbatim}
\end{minipage}
\end{center}
\caption{\label{addvio}Macro to introduce additional constraint
  violation marks.}
\end{figure*}

The application of an OT constraint can now be defined as follows,
using an alternative definition of the optimality operator:

\begin{verbatim}
macro(Cands oo Constraint, 
          Cands
            o
   mark_violation(Constraint)
            o
     ~ range(Cands
               o
         mark_violation(Constraint)
               o
          add_violation)
            o
      {(@ x []),(? -  @)}* ).
\end{verbatim}

Note that this simple approach only works in cases where constraint
violations line up neatly. It turns out that for the syllabification
example discussed earlier that this is the case. Using the {\tt
syllabify} macro given above with this  matching
implementation of the optimality operator produces a transducer of
only 22 states, and can be shown to be exact for all inputs!

\subsection{Permutation}
In the general case, however, constraint violations need not line
up. For example, if the order of constraints is somewhat rearranged
as in: 

\begin{verbatim}
parse oo fill_ons oo have_ons 
    oo fill_nuc oo no_coda
\end{verbatim}

\noindent the matching approach is not exact: it will
produce wrong results for an input such as `arts':

\begin{verbatim}
N[a]D[r]O[t]N[]D[s]      %cf: art@s
N[a]O[r]N[]D[t]O[s]N[]   %cf: ar@ts@
\end{verbatim}
Here, the second output should not be produced because it contains one
more violation of the {\tt fill\_nuc} constraint.  In such cases, a
limited amount of permutation can be used in the filter to make the
marker symbols line up. The {\tt add\_violation} filter of
\fref{addvio} can be extended with the following transducer which
permutes marker symbols:
\begin{verbatim}
macro(permute_marker,
 [{[? *,(@ x []),? *,([] x @)],
   [? *,([] x @),? *,(@ x [])]}*,? *]).
\end{verbatim}
Greater degrees of permutation can be achieved by composing
{\tt permute\_marker} several times. For example:\footnote{An
alternative approach would be to  
compose the {\tt permute\_marker} transducers before inserting extra
markers. Our tests, however, show this alternative to be somewhat less 
efficient.}
\begin{verbatim}
macro(add_violation(3),
 {(bracket x []), (? - bracket)}*   
                o
      [[? *,([] x @)]+, ? *]      
                o
        permute_marker          
                o
        permute_marker
                o
        permute_marker
                o
 {([] x bracket), (? - bracket)}*  ).
\end{verbatim}
So we can incorporate a notion of `precision' in the definition of the
optimality operator for the matching approach as
well, by defining:

\begin{verbatim}
macro(Cands oo Prec :: Constraint),
        Cands
          o
 mark_violation(Constraint)
          o
   ~ range(Cands
             o
       mark_violation(Constraint)
             o
        add_violation(Prec))
          o
   { (@ x []),(? - @)}*  ).
\end{verbatim}

The use of permutation is most effective when constraint violations in 
alternative candidates tend to occur in corresponding positions. In
the worst case, none of the violations may line up. Suppose that for
some constraint, the input ``bebop'' is marked up as:
{\small\begin{verbatim}
c1:   @ b @ e    b    o    p
c2:     b   e  @ b  @ o  @ p
\end{verbatim}}
\noindent In this case, the precision needs to be two in order for
the markers in c1 to line up with markers in c2. Similarly, the
counting approach also needs a precision of two in order to count the
two markers in c1 and prefer this over the greater than two markers in
c2. The general pattern is that any constraint that can be treated
exactly with counting precision N, can also be handled by matching
with precision less than or equal to N. In the other direction,
however, there are constraints, such as those in the Prince and
Smolensky syllabification problem, that can only be exactly
implemented by the matching approach.

For each of the constraint orderings discussed by Prince and
Smolensky, it turns out that at most a single step of permutation
(i.e. a precision of 1) is required for an exact implementation. We
conclude that this OT analysis of syllabification is regular.
This improves upon the result of \newcite{karttunen:98}.
Moreover, the resulting transducers are typically much smaller
too. In \sref{comparison} we present a number of experiments which
provide evidence for this observation. 

\subsection{Discussion}
\paragraph{Containment.}
It might be objected that the Prince and Smolensky syllable structure
example is a particularly simple {\it containment theory} analysis and
that other varieties of OT such as {\it correspondence theory}
\cite{mccarthy-prince} are beyond the scope of
matching.\footnote{\newcite{kager99} compares containment theory and
correspondence theory for the syllable structure example.} Indeed we
have relied on the fact that {\em Gen} only adds brackets and does not add
or delete anything from the set of input symbols. The filter that we
construct needs to compare candidates with alternative candidates
generated {\it from the same input}. 

If {\em Gen} is allowed to change the input then a way must be found to
remember the original input. Correspondence theory is beyond the scope
of this paper, however a simple example of an OT where {\em Gen} modifies
the input is provided by the problem described in \sref{fsot} (from
\newcite{frank-satta}). Suppose we modify {\em Gen} here so that its output
includes a representation of the original input. One way to do this
would be to adopt the convention that input symbols are marked with a
following 0 and output symbols are marked with a following 1. With
this convention {\em Gen} becomes:

\begin{verbatim}
macro(gen, 
 {[(a x [a,0,b,1])*,(b x [b,0,a,1])*], 
  [(a x [a,0,a,1])*,(b x [b,0,b,1])*]})
\end{verbatim}

Then the constraint against the symbol {\tt a} needs to be recast as a
constraint against {\tt [a,1]}.\footnote{OT makes a fundamental
distinction between {\em markedness} constraints (referring only to
the surface) and {\em faithfulness} constraints (referring to both
surface and underlying form). With this mark-up convention,
faithfulness constraints might be allowed to refer to both symbols
marked with 0 and symbols marked with 1. But note that the {\tt Fill}
and {\tt Parse} constraints in syllabification are also considered to 
be faithfulness constraints since they correspond to epenthesis and
deletion respectively.} And, whereas above {\em
add\_violation} was previously written to ignore brackets, for this
case it will need to ignore output symbols (marked with a 1). This
approach is easily implementable and with sufficient use of
permutation, an approximation can be achieved for any predetermined
bound on input length.

\paragraph{Locality.}
In discussing the impact of their result, \newcite{frank-satta}
suggest that the OT formal system is too rich in generative capacity.
They suggest a {\em shift in the type of optimization carried out in
  OT, from global optimization over arbitrarily large representations
  to local optimization over structural domains of bounded
  complexity}. The approach of matching constraint violations proposed
here is based on the assumption that constraint violations can indeed
be compared {\em locally}.  

However, if {\em locality} is crucial then one might wonder why we
extended the local matching approach with global permutation steps.
Our motivation for the use of global permutation is the observation
that it ensures the matching approach is strictly more powerful than
the counting approach. A weaker, and perhaps more interesting,
treatment is obtained if locality is enforced in these permutation
steps as well. For example, such a weaker variant is obtained if the
following definition of {\tt permute\_marker} is used:

\begin{verbatim}
macro(permute_marker,   % local variant
 {? ,[([] x @),?,(@ x [])],
     [(@ x []),?,([] x @)]}* ).
\end{verbatim}
This is a weaker notion of permutation than the definition given
earlier. Interestingly, using this definition resulted in equivalent
transducers for all of the syllabification examples given in this
paper. In the general case, however, matching with local permutation is less
powerful. 

\begin{figure}
\includegraphics[width=.45\textwidth]{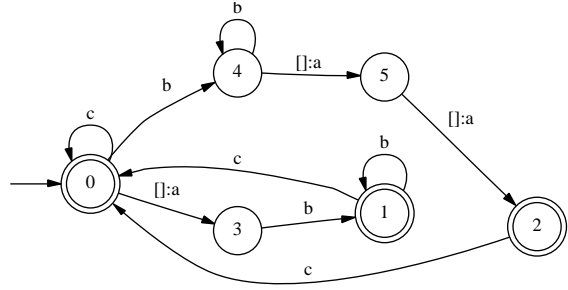}
\centering
\caption{\label{genfig}{\em Gen} for an example for which local permutation
  is not sufficient.}
\end{figure}

Consider the following artificial example. In this example, inputs of
interest are strings over the alphabet $\{b,c\}$. {\em Gen} introduces
an {\tt a} {\em before} a sequence of {\tt b}'s, or two {\tt a}'s {\em
after} a sequence of {\tt b}'s. {\em Gen} is given as an automaton in
\fref{genfig}.  There is only a single constraint, which forbids {\tt
a}. It can easily be verified that a matching approach with global
permutation using a precision of 1 exactly implements this OT. In
contrast, both the counting approach as well as a matching approach
based on local permutation can only approximate this
OT.\footnote{Matching with local permutation is not strictly more
powerful than counting. For an example, change {\em Gen} in this
example to: {\tt \{[([] x a),\{b,c\}*],[\{b,c\}*,([] x
[a,a])]\}}. This can be exactly implemented by counting with a
precision of one. Matching with local permutation, however, cannot
exactly implement this case, since markers would need to be permuted
across unbounded sequences.}

\section{\label{comparison}Comparison}

In this section we compare the two alternative approaches with respect
to accuracy and the number of states of the resulting transducers.
We distinguish between {\em exact} and {\em approximating}
implementations. An implementation is exact if it produces the right
result for all possible inputs. 

Assume we have a transducer $T$ which correctly implements an OT
analysis, except that it perhaps fails to distinguish between
different numbers of constraint violations for one or more relevant
constraints. We can decide whether this $T$ is exact as follows. $T$
is exact if and only if $T$ is exact with respect to each of the
relevant constraints, i.e., for each constraint, $T$ distinguishes
between different numbers of constraint violations. In order to check
whether $T$ is exact in this sense for constraint $C$ we create the
transducer {\tt is\_exact(T,C)}:
\begin{verbatim}
macro(is_exact(T,C),
           T 
           o 
     mark_violation(C) 
           o 
   {(? - @) x [], @}*).
\end{verbatim}
If there are inputs for which this transducer produces multiple
outputs, then we know that {\tt T} is not exact for {\tt C}; otherwise
{\tt T} {\em is} exact for {\tt C}.  This reduces to the question
of whether {\tt is\_exact(T,C)} is ambiguous. The question of whether a
given transducer is ambiguous is shown to be decidable in
\cite{blattner-head}; and an efficient algorithm is proposed in
\cite{roch:lang97}.\footnote{We have adapted the algorithm proposed in
  \cite{roch:lang97} since it fails to treat certain types of
  transducer correctly; we intend to provide details somewhere else. }
Therefore, in order to check a given transducer {\tt T} for exactness,
it must be the case that for each of the constraints {\tt C}, {\tt
  is\_exact(T,C)} is nonambiguous.

If a transducer {\tt T} is not exact, we characterize the quality of the
approximation by considering the maximum length of input strings for
which {\tt T} is exact. For example, even though {\tt T} fails the
exactness check, it might be the case that 
\begin{verbatim}
    [? ^,? ^,? ^,? ^,? ^] 
            o 
            T
\end{verbatim}
in fact is exact, indicating that {\tt T} produces the correct result
for all inputs of length $\leq 5$.

Suppose we are given the sequence of constraints:
\begin{verbatim}
 have_ons >> fill_ons >> parse 
    >> fill_nuc >> no_coda
\end{verbatim}
and suppose furthermore that we require that the implementation, using
the counting approach, must
be exact for all strings of length $\leq 10$. How can we determine the
level of precision for each of the constraints?  A simple algorithm
(which does not necessarily produce the smallest transducer) proceeds
as follows. Firstly, we determine the precision of the first, most
important, constraint by checking exactness for the transducer 
\begin{verbatim}
gen oo P :: have_ons
\end{verbatim}
for increasing values for {\tt P}. As soon as we find the minimal {\tt
  P} for which the exactness check succeeds (in this case for P=0), we
continue by determining the precision required for the next constraint
by finding the minimal value of P in:
\begin{verbatim}
gen oo 0 :: have_ons oo P :: fill_ons
\end{verbatim}
We continue in this way until we have determined precision values for
each of the constraints. In this case we obtain a transducer with 8269
states implementing:
\begin{verbatim}
gen oo 0 :: have_ons
    oo 1 :: fill_ons
    oo 8 :: parse
    oo 5 :: fill_nuc
    oo 4 :: no_coda
\end{verbatim}
In contrast, using matching an exact implementation is obtained using
a precision of 1 for the {\tt fill\_nuc} constraint; all other
constraints have a precision of 0. This transducer contains only 28
states.  

The assumption in OT is that each of the constraints is universal, 
whereas the constraint {\em order} differs from language to
language. Prince and Smolensky identify nine interestingly different
constraint orderings. These nine ``languages'' are presented in
table~\ref{langs}.

\begin{table*}
\begin{center}
\begin{tabular}{cc}
id & constraint order \\
1 & \tt have\_ons $\gg$ fill\_ons $\gg$ no\_coda  $\gg$ fill\_nuc $\gg$ parse    \\
2 & \tt have\_ons $\gg$ no\_coda  $\gg$ fill\_nuc $\gg$ parse    $\gg$ fill\_ons \\
3 & \tt no\_coda  $\gg$ fill\_nuc $\gg$ parse    $\gg$ fill\_ons $\gg$ have\_ons \\
4 & \tt have\_ons $\gg$ fill\_ons $\gg$ no\_coda  $\gg$ parse    $\gg$ fill\_nuc \\
5 & \tt have\_ons $\gg$ no\_coda  $\gg$ parse    $\gg$ fill\_nuc $\gg$ fill\_ons \\
6 & \tt no\_coda  $\gg$ parse    $\gg$ fill\_nuc $\gg$ fill\_ons $\gg$ have\_ons \\
7 & \tt have\_ons $\gg$ fill\_ons $\gg$ parse    $\gg$ fill\_nuc $\gg$ no\_coda  \\
8 & \tt have\_ons $\gg$ parse    $\gg$ fill\_ons $\gg$ fill\_nuc $\gg$ no\_coda  \\
9 & \tt parse    $\gg$ fill\_ons $\gg$ have\_ons $\gg$ fill\_nuc $\gg$ no\_coda  \\
\end{tabular}
\end{center}
\caption{\label{langs}Nine different constraint orderings for
  syllabification, as given in Prince and Smolensky, chapter 6.}
\end{table*}

In table~\ref{comp} we compare the size of the resulting automata for
the matching approach, as well as for the counting approach, for three
different variants which are created in order to guarantee exactness
for strings of length $\leq 5$, $\leq 10$ and $\leq 15$ respectively.

\begin{table*}
\begin{center}
\begin{tabular}{|cc|rrrrrrrrr|}
\hline
Method &Exactness   & \multicolumn{9}{c|}{Constraint order}                      \\
          &         &    1 &   2 &    3 &    4 &  5 &   6 &    7 &     8 &     9\\ 
\hline
matching & exact     &   29 &  22 &   20 &   17 & 10 &   8 &   28 &    23 &    20\\
\hline
counting & $\leq 5$  &   95 & 220 &  422 &  167 & 10 & 240 & 1169 &  2900 &  4567\\
counting & $\leq 10$ &  280 & 470 & 1667 &  342 & 10 & 420 & 8269 & 13247 & 16777\\
counting & $\leq 15$ &  465 & 720 & 3812 &  517 & 10 & 600 &22634 & 43820 & 50502\\
\hline
\end{tabular}
\end{center}

\caption{\label{comp}Comparison of the matching approach and the
  counting approach for various levels of exactness. The numbers
  indicate the number of states of the resulting transducer. }
\end{table*}

Finally, the construction of the transducer using the matching
approach is typically much faster as well. In table~\ref{timings} some
comparisons are summarized.

\begin{table*}
\begin{center}
\begin{tabular}{|cc|rrrrrrrrr|}
\hline
Method  & Exactness &\multicolumn{9}{c|}{Constraint order}                \\
        &           &   1 &   2 &    3 &   4 &  5 &  6 &   7 &    8 &    9\\ 
\hline
matching& exact     & 1.0 & 0.9 &  0.9 & 0.9 &0.8 &0.7 & 1.5 &  1.3 &  1.1\\
\hline
counting& $\leq  5$ & 0.9 & 1.7 &  4.8 & 1.6 &0.5 &1.9 &10.6 & 18.0 & 30.8\\
counting& $\leq 10$ & 2.8 & 4.7 & 28.6 & 4.0 &0.5 &4.2 &83.2 &112.7 &160.7\\
counting& $\leq 15$ & 6.8 &10.1 & 99.9 & 8.6 &0.5 &8.2 &336.1&569.1 &757.2\\
\hline
\end{tabular}
\end{center}

\caption{\label{timings}Comparison of the matching approach and the
  counting approach for various levels of exactness. The numbers
  indicate the CPU-time in seconds required to construct the transducer. }
\end{table*}

\section{Conclusion}

We have presented a new approach for implementing OT which is based on
matching rather than the counting approach of \newcite{karttunen:98}. The
matching approach shares the advantages of the counting approach in
that it uses the finite state calculus and avoids off-line sorting and
counting of constraint violations. We have shown that the matching
approach is superior in that analyses that can only be approximated
by counting can be exactly implemented by matching. Moreover, the
size of the resulting transducers is significantly smaller.

We have shown that the matching approach along with global permutation
provides a powerful technique technique for minimizing constraint
violations. Although we have only applied this approach to
permutations of the Prince \& Smolensky syllabification analysis, we
speculate that the approach (even with local permutation) will also
yield exact implementations for most other OT phonological
analyses. Further investigation is needed here, particularly with
recent versions of OT such as correspondence theory. Another line of
further research will be the proper integration of finite state OT
with non-OT phonological rules as discussed, for example, in papers
collected in \newcite{hermans99} . 

Finally, we intend also to investigate the application of our approach
to syntax. \newcite{karttunen:98} suggests that the Constraint Grammar
approach of \newcite{karlsson95} could be implemented using lenient
composition. If this is the case, it could most probably be
implemented more precisely using the matching approach. Recently,
\newcite{oflazer99} has presented an implementation of Dependency syntax
which also uses lenient composition with the counting approach. The
alternative of using a matching approach here should be
investigated.

\bibliographystyle{acl}

\begin{thebibliography}{}

\bibitem[\protect\citename{Blattner and Head}1977]{blattner-head}
Meera Blattner and Tom Head.
\newblock 1977.
\newblock Single-valued a-transducers.
\newblock {\em Journal of Computer and System Sciences}, 15(3):328--353.

\bibitem[\protect\citename{Ellison}1994]{ellison}
Mark~T. Ellison.
\newblock 1994.
\newblock Phonological derivation in optimality theory.
\newblock In {\em Proceedings of the 15th International Conference on
  Computational Linguistics (COLING)}, pages 1007--1013, Kyoto.

\bibitem[\protect\citename{Frank and Satta}1998]{frank-satta}
Robert Frank and Giorgio Satta.
\newblock 1998.
\newblock Optimality theory and the computational complexity of constraint
  violability.
\newblock {\em Computational Linguistics}, 24:307--315.

\bibitem[\protect\citename{Gerdemann and van
  Noord}1999]{eacl99-gerdemann-vannoord}
Dale Gerdemann and Gertjan van Noord.
\newblock 1999.
\newblock Transducers from rewrite rules with backreferences.
\newblock In {\em Ninth Conference of the European Chapter of the Association
  for Computational Linguistics}, Bergen Norway.

\bibitem[\protect\citename{Hermans and van Oostendorp}1999]{hermans99}
Ben Hermans and Marc van Oostendorp, editors.
\newblock 1999.
\newblock {\em The Derivational Residue in Phonological Optimality Theory},
  volume~28 of {\em Linguistik Aktuell/Linguistics Today}.
\newblock John Benjamins, Amsterdam/Philadelphia.

\bibitem[\protect\citename{Johnson}1972]{john:form72}
C.~Douglas Johnson.
\newblock 1972.
\newblock {\em Formal Aspects of Phonological Descriptions}.
\newblock Mouton, The Hague.

\bibitem[\protect\citename{Kager}1999]{kager99}
Ren\'e Kager.
\newblock 1999.
\newblock {\em Optimality Theory}.
\newblock Cambridge UP, Cambridge, UK.

\bibitem[\protect\citename{Kaplan and Kay}1994]{kapl:regu94}
Ronald Kaplan and Martin Kay.
\newblock 1994.
\newblock Regular models of phonological rule systems.
\newblock {\em Computational Linguistics}, 20(3):331--379.

\bibitem[\protect\citename{Karlsson \bgroup et al.\egroup }1995]{karlsson95}
Fred Karlsson, Atro Voutilainen, Juha Heikkil\"a, and Arto Anttila.
\newblock 1995.
\newblock {\em Constraint Grammar: A Language-Independent Framework for Parsing
  Unrestricted Text}.
\newblock Mouton de Gruyter, Berlin/New York.

\bibitem[\protect\citename{Karttunen \bgroup et al.\egroup }1996]{kart:regu96}
Lauri Karttunen, Jean-Pierre Chanod, Gregory Grefenstette, and Anne Schiller.
\newblock 1996.
\newblock Regular expressions for language engineering.
\newblock {\em Natural Language Engineering}, 2(4):305--238.

\bibitem[\protect\citename{Karttunen}1995]{kart:95}
Lauri Karttunen.
\newblock 1995.
\newblock The replace operator.
\newblock In {\em 33th Annual Meeting of the Association for Computational
  Linguistics}, M.I.T. Cambridge Mass.

\bibitem[\protect\citename{Karttunen}1996]{karttunen:96}
Lauri Karttunen.
\newblock 1996.
\newblock Directed replacement.
\newblock In {\em 34th Annual Meeting of the Association for Computational
  Linguistics}, Santa Cruz.

\bibitem[\protect\citename{Karttunen}1998]{karttunen:98}
Lauri Karttunen.
\newblock 1998.
\newblock The proper treatment of optimality theory in computational phonology.
\newblock In {\em Finite-state Methods in Natural Language Processing}, pages
  1--12, Ankara.

\bibitem[\protect\citename{Kempe and Karttunen}1996]{KempeKarttunen}
Andr\'e Kempe and Lauri Karttunen.
\newblock 1996.
\newblock Parallel replacement in the finite-state calculus.
\newblock In {\em Proceedings of the 16th International Conference on
  Computational Linguistics (COLING)}, Copenhagen, Denmark.

\bibitem[\protect\citename{Kisseberth}1970]{kisseberth}
Charles Kisseberth.
\newblock 1970.
\newblock On the functional unity of phonological rules.
\newblock {\em Linguistic Inquiry}, 1:291--306.

\bibitem[\protect\citename{Mc{C}arthy and Prince}1995]{mccarthy-prince}
John Mc{C}arthy and Alan Prince.
\newblock 1995.
\newblock Faithfulness and reduplicative identity.
\newblock In Jill Beckman, Laura~Walsh Dickey, and Suzanne Urbanczyk, editors,
  {\em Papers in Optimality Theory}, pages 249--384. Graduate Linguistic
  Student Association, Amherst, Mass.
\newblock University of Massachusetts Occasional Papers in Linguistics 18.

\bibitem[\protect\citename{Mohri and Sproat}1996]{mohri-sproat:96}
Mehryar Mohri and Richard Sproat.
\newblock 1996.
\newblock An efficient compiler for weighted rewrite rules.
\newblock In {\em 34th Annual Meeting of the Association for Computational
  Linguistics}, Santa Cruz.

\bibitem[\protect\citename{Oflazer}1999]{oflazer99}
Kemal Oflazer.
\newblock 1999.
\newblock Dependency parsing with an extended finite state approach.
\newblock In {\em 37th Annual Meeting of the Association for Computational
  Linguistics}, pages 254--260.

\bibitem[\protect\citename{Prince and Smolensky}1993]{prince-smolensky:93}
Alan Prince and Paul Smolensky.
\newblock 1993.
\newblock Optimality theory: Constraint interaction in generative grammar.
\newblock Technical Report TR-2, Rutgers University Cognitive Science Center,
  New Brunswick, NJ.
\newblock MIT Press, To Appear.

\bibitem[\protect\citename{Roche and Schabes}1997]{roch:lang97}
Emmanuel Roche and Yves Schabes.
\newblock 1997.
\newblock Introduction.
\newblock In Emmanuel Roche and Yves Schabes, editors, {\em Finite-State
  Language Processing}. MIT Press, Cambridge, Mass.

\bibitem[\protect\citename{Tesar}1995]{tesar}
Bruce Tesar.
\newblock 1995.
\newblock {\em Computational Optimality Theory}.
\newblock {Ph.D.} thesis, University of Colorado, Boulder.

\bibitem[\protect\citename{van Noord and Gerdemann}1999]{wia99}
Gertjan van Noord and Dale Gerdemann.
\newblock 1999.
\newblock An extendible regular expression compiler for finite-state approaches
  in natural language processing.
\newblock In O.~Boldt, H.~Juergensen, and L.~Robbins, editors, {\em Workshop on
  Implementing Automata; WIA99 Pre-Proceedings}, Potsdam, Germany.

\bibitem[\protect\citename{van Noord}1997]{fsa-two}
Gertjan van Noord.
\newblock 1997.
\newblock {FSA Utilities}: {A} toolbox to manipulate finite-state automata.
\newblock In Darrell Raymond, Derick Wood, and Sheng Yu, editors, {\em Automata
  Implementation}, pages 87--108. Springer Verlag.
\newblock Lecture Notes in Computer Science 1260.

\bibitem[\protect\citename{van Noord}1999]{noord:fsa6}
Gertjan van Noord.
\newblock 1999.
\newblock {FSA6} reference manual.
\newblock The {\em FSA Utilities\/} toolbox is available free of charge under
  Gnu General Public License at http://www.let.rug.nl/\~{}vannoord/Fsa/.

\bibitem[\protect\citename{Walther}1996]{walther}
Markus Walther.
\newblock 1996.
\newblock {OT} simple -- a construction-kit approach to optimality theory
  implementation.
\newblock ROA-152-1096.

\end{thebibliography}

\end{document}